\pdfoutput=1
\documentclass[letterpaper, 10 pt, conference]{cls/ieeeconf}

\IEEEoverridecommandlockouts  
\let\labelindent\relax

\usepackage[table,usenames,dvipsnames]{xcolor}
\usepackage{cite}
\usepackage{amsmath,amssymb,amsfonts,amsthm,amscd,dsfont}
\usepackage{accents} 
\usepackage{bbm}
\usepackage{algorithm,listings}
\usepackage[noend]{algpseudocode}
\usepackage{graphicx,tabularx,adjustbox}
\usepackage[font={small}]{caption}
\usepackage{subcaption}
\usepackage{paralist}

\usepackage[titletoc,title]{appendix}
\usepackage{balance}
\usepackage[breaklinks=true, colorlinks, bookmarks=true, citecolor=Black, urlcolor=Violet,linkcolor=Black]{hyperref}

\newcommand{\calC}{{\cal C}}

\newcommand{\calE}{{\cal E}}

\newcommand{\calG}{{\cal G}}

\newcommand{\calN}{{\cal N}}

\newcommand{\calP}{{\cal P}}

\newcommand{\calS}{{\cal S}}

\newcommand{\calV}{{\cal V}}

\newcommand{\bfe}{\mathbf{e}}

\newcommand{\bfg}{\mathbf{g}}
\newcommand{\bfh}{\mathbf{h}}

\newcommand{\bfm}{\mathbf{m}}

\newcommand{\bfq}{\mathbf{q}}

\newcommand{\bfalpha}{\boldsymbol{\alpha}}
\newcommand{\bfbeta}{\boldsymbol{\beta}}

\newcommand{\bbE}{\mathbb{E}}

\newcommand{\bbR}{\mathbb{R}}

\newcommand{\crl}[1]{\left\{#1\right\}}

\algnewcommand{\LineComment}[1]{\State \(\triangleright\) #1}

\theoremstyle{definition}

\newtheorem*{problem}{Problem}
\theoremstyle{remark}

\title{\LARGE \bf Distributed Optimization with Consensus Constraint for Multi-Robot Semantic Octree Mapping%
\thanks{}%
}

\author{Arash Asgharivaskasi \and Nikolay Atanasov
\thanks{The authors are with the Department of Electrical and Computer Engineering, University of California San Diego, CA 92093, USA {\tt\small \{aasghari,natanasov\}@eng.ucsd.edu}.}%
\thanks{We gratefully acknowledge support from ARL DCIST CRA W911NF-17-2-0181 and NSF FRR CAREER 2045945.}
}

\begin{document}

\maketitle

\begin{abstract}
This work develops a distributed optimization algorithm for multi-robot 3-D semantic mapping using streaming range and visual observations and single-hop communication. Our approach relies on gradient-based optimization of the observation log-likelihood of each robot subject to a map consensus constraint to build a common multi-class map of the environment. This formulation leads to closed-form updates which resemble Bayes rule with one-hop prior averaging. To reduce the amount of information exchanged among the robots, we utilize an octree data structure that compresses the multi-class map distribution using adaptive-resolution. 
\end{abstract}

\section{Introduction}
\label{sec:intro}

Dense semantically-rich environment representations, such as multi-class volumetric maps~\cite{vaskasi_TRO}, can be built online by mobile robots thanks to the availability of on-board GPU-accelerated segmentation models. Distributed 3-D mapping requires designing optimization algorithms that take into account the communication and computation limits of multi-robot systems. Among the recent works in distributed mapping for robotics, Corah \textit{et al.} \cite{corah} proposed a distributed Gaussian mixture model (GMM) in which a GMM representation is communicated over a network of robots and is utilized by each robot to build a local occupancy grid map for planning. The distributed estimation work by Paritosh \textit{et al.} \cite{Paritosh_dist_mapping} uses mirror descent optimization of an information-theoretic cost function to obtain the probability density function (PDF) of unknown continuous variables by communicating local estimates among the agents. The metric-semantic mapping approach in Kimera-multi~\cite{kimera-multi} builds a dense multi-class mesh representation from visual-inertial observations and local trajectory communication used to estimate a global reference frame in a decentralized manner.

In this work, we propose a multi-class mapping algorithm for a team of mobile robots. Our main \textbf{contribution} is an online distributed multi-robot semantic octree mapping method, where each node in the octree maintains a categorical distribution over various object classes, and is updated using local range and category observations. We approach the multi-robot mapping problem using distributed gradient optimization of the observation log-likelihood, and show that every iteration contains a consensus step where neighbors average their map distributions and a Bayesian update step where new local observations are incorporated into the map distribution by each robot. Through a simulated experiment, we show that our algorithm leads to globally consistent Bayesian multi-class mapping and the octree data structure helps to achieve memory and communication efficiency.

\section{Problem Statement}
\label{sec:prob_statement}

\begin{figure}[t]
    \centering
    \includegraphics[width=0.9\linewidth]{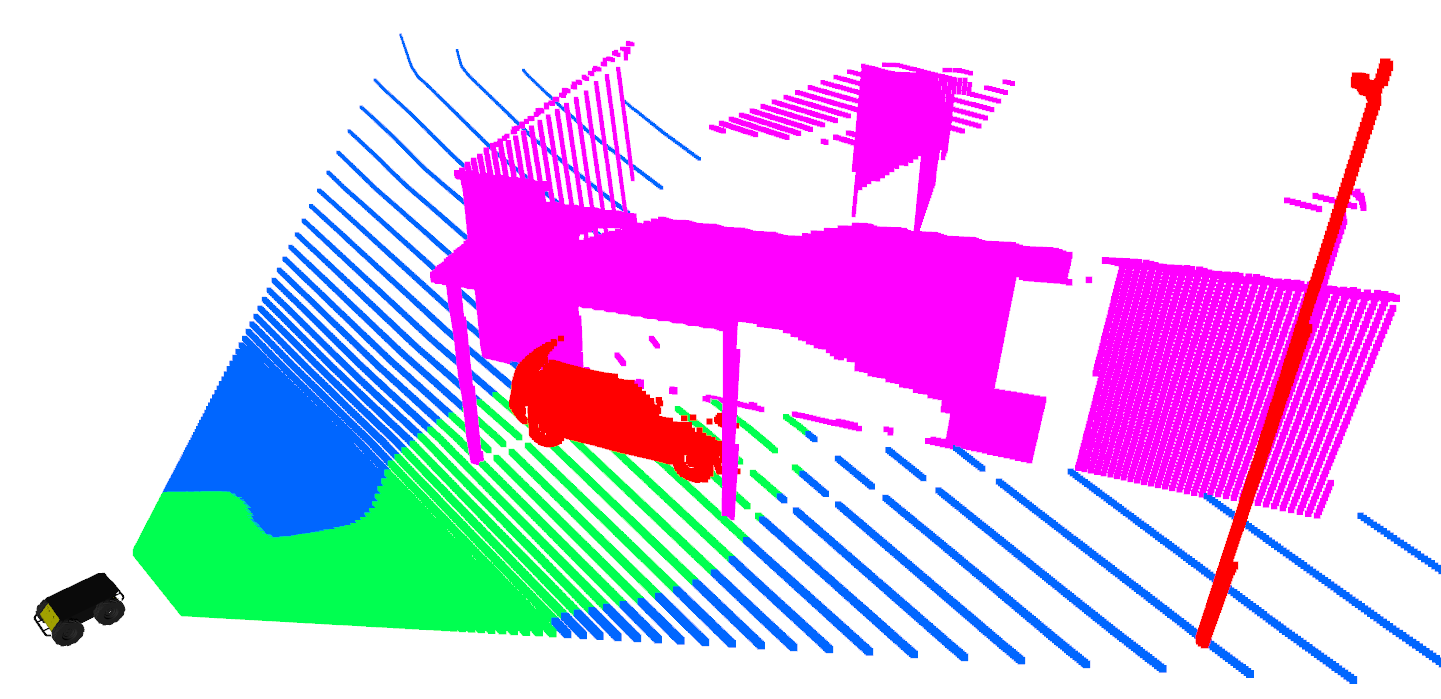}\\
    \caption{Semantically annotated point cloud obtained from a range and category observation $z_t^i$. Each object category is shown with a unique color.}
    \label{fig:setup}
\end{figure}

Consider a network of robots represented by an undirected graph $\calG(\calV, \calE)$ where $\calV$ denotes the set of robots and $\calE \subseteq \calV \times \calV$ encodes the existence of a communication link between each pair of robots, assumed to be constant throughout the task. The robots are navigating in an environment consisting of a collection of disjoint sets $\calS_k \subset \mathbb{R}^3$, each associated with a semantic category $c \in \calC := \crl{0,1,\ldots,C}$. Let $\calS_0$ denote free space, while each $\calS_c$ for $c >0$ represents a different category, such as building, vegetation, terrain. Each robot $i$ is equipped with a sensor that provides a stream of ray-based observations $z^i_t \in \bbR \times \calC$ at time $t$. Each observation contains information about the distance to and semantic category of the observed object along the ray. See Fig.~\ref{fig:setup} for an example. Such information may be obtained by processing the measurements of an RGBD camera or a LiDAR with a semantic segmentation algorithm. 

The robots aim to construct a common multi-class map $\bfm$ of the environment by utilizing their semantically-annotated local range measurements as well as communication with their teammates. We model the map $\bfm$ as a grid of independent cells, where each individual cell $m$ is labeled with a category in $\calC$. To model measurement noise, we consider a PDF $p(z^i_t \mid \bfm)$ for each observation. The goal is to compute the map estimate that maximizes the sum of expected log-likelihood of data, i.e. local observations $\{z_{\tau}^i\}_{\tau=1}^t$ for each robot $i \in \calV$:
\begin{equation}
    \max_{p \in \calP} \sum_{i=1}^N \bbE_{\bfm \sim p} \big[\sum_{\tau = 1}^t \log{p(z_{\tau}^i|\bfm)}\big],\nonumber
\end{equation}
where $N = |\calV|$ and $\calP$ is the space of all probability mass functions (PMF) over the set of possible maps. To remove the constraint $p \in \calP$, we utilize a multi-class log-odds ratio representation of the categorical distribution \cite{vaskasi_TRO}:
\begin{equation*}
\bfh := \begin{bmatrix} \log \frac{p(m = 0)}{p(m = 0)} & \cdots & \log \frac{p(m = C)}{p(m = 0)} \end{bmatrix}^\top \in \bbR^{C+1}.
\end{equation*}
A probability mass function and its log-odds representation have a one-to-one correspondence through the softmax function $\sigma:\mathbb{R}^{C+1} \mapsto \mathbb{R}^{C+1}$:
\begin{equation}
p(m = c) = \sigma_{c+1}(\bfh) := \frac{\bfe_{c+1}^\top \exp(\bfh)}{\mathbf{1}^\top \exp(\bfh)},\nonumber
\end{equation}
where $\bfe_c$ is the standard basis vector with $c$-th element equal to $1$ and $0$ elsewhere, $\mathbf{1}$ is the vector with all elements equal to $1$, and $\exp(\cdot)$ is applied element-wise to the vector $\bfh$. Applying Bayes rule to the observation model $p(z^i_{\tau}|\bfm)$ and using the map cell independence assumption makes it possible to solve the optimization for individual cells instead of maintaining the joint distribution over all cells. Thus, the estimation objective can then be written as an unconstrained optimization for each map cell:
\begin{gather}
    \max_{\bfh \in \bbR^{C+1}} \sum_{i=1}^N \sum_{m=0}^C \sigma_{m+1}(\bfh) \log{\frac{q_i(m)}{\sigma_{m+1}(\bfh)}},\nonumber
\end{gather}
where $q_i(m) = \sqrt[t]{\prod_{\tau=1}^t p(m | z_{\tau}^i)}$ is the temporal geometric average of the inverse observation model. In order to enable distributed optimization of the objective, we introduce a constraint that requires the robots to agree on a common map estimate via only one-hop communication as follows.

\begin{problem}
    Let $\calG(\calV, \calE)$ be a network of $N$ robots, where each robot $i \in \calV$ collects observations $\{z_{\tau}^i\}_{\tau=1}^t$ and maintains its own \textit{local estimate} of the map cell distribution $\bfh_i$. Estimate the globally consistent common map such that:
    \begin{align}
    \max_{\bfh_{1:N} \in \bbR^{(C+1) \times N}} &\sum_{i=1}^N \sum_{m=0}^C \sigma_{m+1}(\bfh_i) \log{\frac{q_i(m)}{\sigma_{m+1}(\bfh_i)}},\nonumber\\
    \text{s.t.} \quad &\sum_{\{i,j\} \in \calE} A_{ij} \|\bfh_{j} - \bfh_{i}\|_2^2 = 0,\label{eq:dist_mapping_log_odds}
    \end{align}
where $A_{ij}$ is the $ij$-th entry of the weighted adjacency matrix of $\calG$ with $\sum_j A_{ij} = 1$.
\end{problem}
\section{Distributed Semantic Mapping}
\label{sec:method}

The optimization objective in \eqref{eq:dist_mapping_log_odds} has a specific structure as the sum of local objective functions over all robots, subject to a consensus constraint among all $\bfh_i$, $i \in \calV$. We develop an iterative gradient-based method inspired by the general distributed optimization framework presented in \cite{shah2017distributed}. The underlying idea is to interleave local gradient updates for the individual objective function and for the consensus constraint at each robot. Alg.~\ref{alg:dist_mapping} presents the update steps for each map cell in order to solve \eqref{eq:dist_mapping_log_odds} in a distributed manner. We initialize the log-odds $\bfh^{(0)}_i$ with prior $\bfh_0$ for each robot $i$. We continuously apply the update steps until a maximum number of iterations is reached, or the update norm is smaller than a threshold. The update step in line~\ref{alg_line:consensus} guides the local log-odds towards satisfaction of the consensus constraint, which only requires single-hop communication between neighboring robots $j \in \calN_i$. Line~\ref{alg_line:grad_comp} incorporates the local observations $\log{\bfq_i} = [\log{q_i(m)}]_{m=0}^{C}$ via $\boldsymbol{\Delta}_i$, where $\odot$ is the element-wise multiplication. This step is only local to each robot $i$ and does not require communication. Note that both update steps \ref{alg_line:consensus} and \ref{alg_line:grad_apply} resemble the log-odds equivalent of the Bayes rule for updating multi-class probabilities (see (8) in \cite{vaskasi_TRO}). Alg.~\ref{alg:dist_mapping} results in a globally consistent multi-class probabilistic grid map of the environment that is shared amongst all robots in $\calV$. Proofs of convergence to consensus and optimality with respect to the objective function \eqref{eq:dist_mapping_log_odds} will be shown in future work.

The distributed semantic mapping algorithm we developed assumes a regular grid representation of the environment. To reduce the storage and communication requirements, we may utilize a semantic octree data structure which provides a lossless compression of the original 3-D multi-class map. In this case, the update rules in Alg.~\ref{alg:dist_mapping} should be applied to all leaf nodes in the semantic octree map of each robot $i$. Please refer to Alg.~3 in \cite{vaskasi_TRO} for the octree equivalents of the update steps in \ref{alg_line:consensus} and \ref{alg_line:grad_apply}.

\begin{algorithm}
    \caption{Distributed Semantic Mapping}
\begin{algorithmic}[1]
\renewcommand{\algorithmicrequire}{\textbf{Input:}}
\renewcommand{\algorithmicensure}{\textbf{Output:}}
\Require Local observations $\{z_{\tau}^i\}_{\tau=1}^t$ for each robot $i \in \calV$
\State $\bfh^{(0)}_i = \bfh_0 \quad \forall i \in \calV$ \Comment{Initialize all log-odds to $\bfh_0$}
\State $k = 0$
\Loop
    \For{each robot $i \in \calV$}
        \LineComment{Enforce consensus with step size $\epsilon$:}
        \State $\Tilde{\bfh}^{(k+1)}_i = \bfh^{(k)}_i + 2 \epsilon \sum_{j \in \calN_i} A_{ij} (\bfh^{(k)}_j - \bfh^{(k)}_i)$ \label{alg_line:consensus}
        \LineComment{Gradient computation:}
        \State $\boldsymbol{\Delta}_i = \Tilde{\bfh}^{(k+1)}_i - \log{\bfq_i}$
        \State $\bfalpha^{(k+1)}_i = (\exp(\Tilde{\bfh}^{(k+1)}_i)^{\top} \boldsymbol{\Delta}_i) \mathbf{1}$
        \State $\bfbeta^{(k+1)}_i = (\exp(\Tilde{\bfh}^{(k+1)}_i)^{\top} \mathbf{1}) \boldsymbol{\Delta}_i$
        \State $\bfg^{(k+1)}_i = (\bfalpha^{(k+1)}_i - \bfbeta^{(k+1)}_i) \odot \frac{\exp(\Tilde{\bfh}^{(k+1)}_i)}{(\exp(\Tilde{\bfh}^{(k+1)}_i)^{\top} \mathbf{1})^2}$ \label{alg_line:grad_comp}
        \LineComment{Apply gradient with step size $\gamma^{(k+1)}$:}
        \State $\bfh^{(k+1)}_i = \Tilde{\bfh}^{(k+1)}_i + \gamma^{(k+1)} \bfg^{(k+1)}_i$ \label{alg_line:grad_apply}
        \State $h^{(k+1)}_{i,1} = 0$ \Comment{$h^{(k+1)}_{i,1} = \log\frac{p(m=0)}{p(m=0)} = 0$}
    \EndFor
    \State $k = k+1$
\EndLoop
\State \Return $\bfh_i^{(k)} \quad \forall i \in \calV$
\end{algorithmic}
\label{alg:dist_mapping}
\end{algorithm}
\section{Evaluation}
\label{sec:eval}

\begin{figure}[t]
    \begin{subfigure}[t]{0.49\linewidth}
    \centering
    \includegraphics[width=\linewidth]{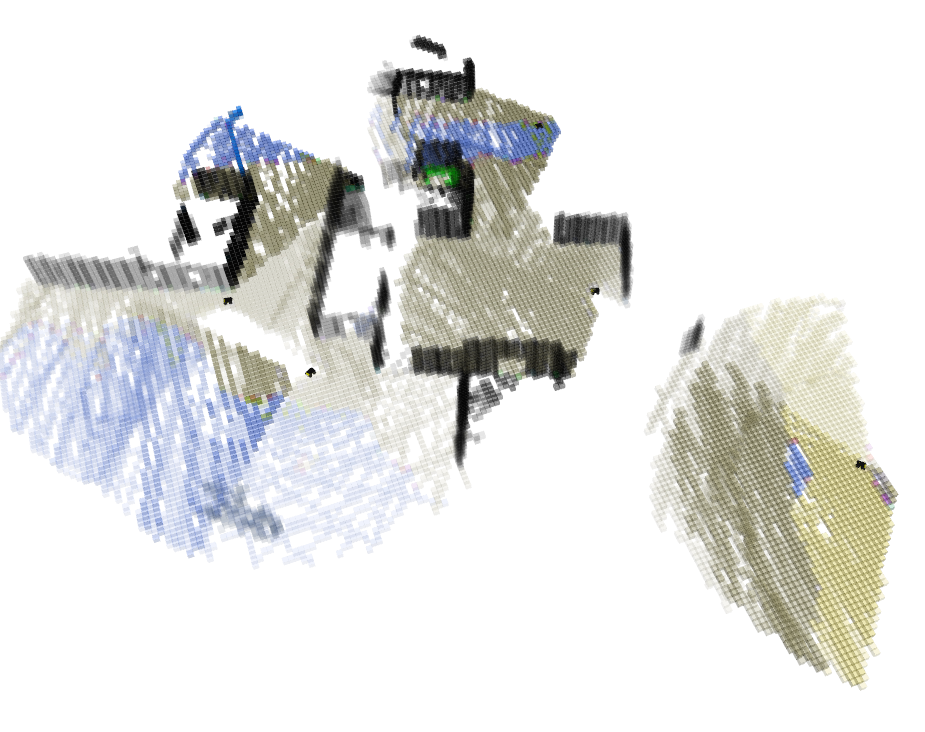}
    \captionsetup{justification=centering}
    \caption{Multi-robot exploration begins}
    \end{subfigure}%
    \hfill%
    \begin{subfigure}[t]{0.49\linewidth}
    \centering
    \includegraphics[width=\linewidth]{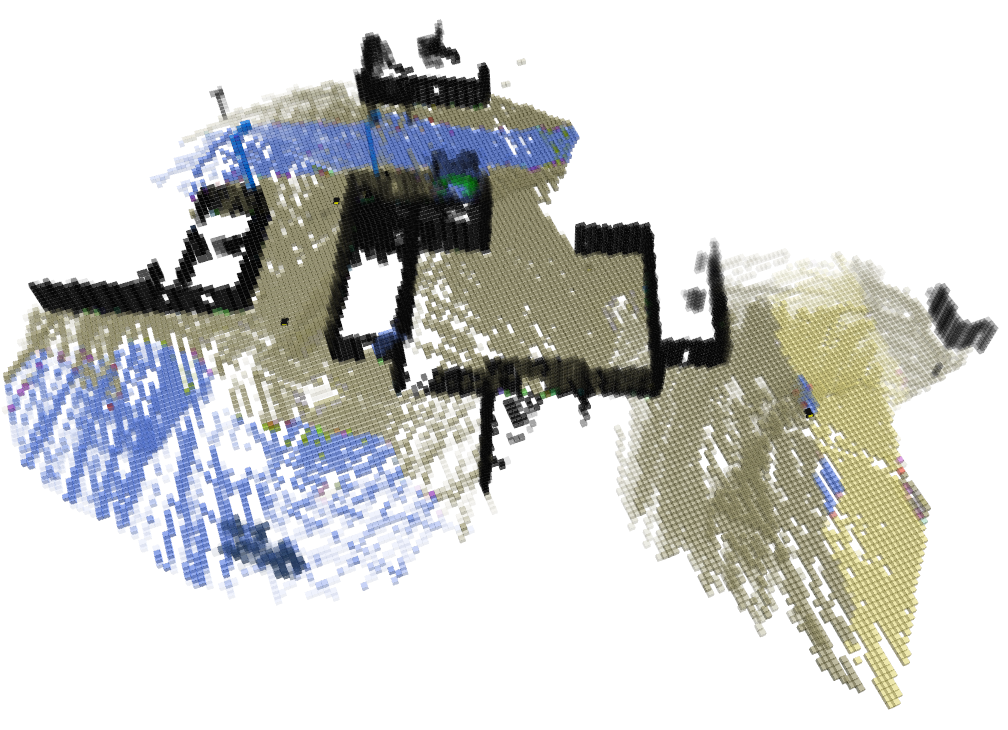}
    \caption{Exploration at $t=100\text{s}$}
    \end{subfigure}\\
    \begin{subfigure}[t]{\linewidth}
    \centering
    \includegraphics[width=0.85\linewidth]{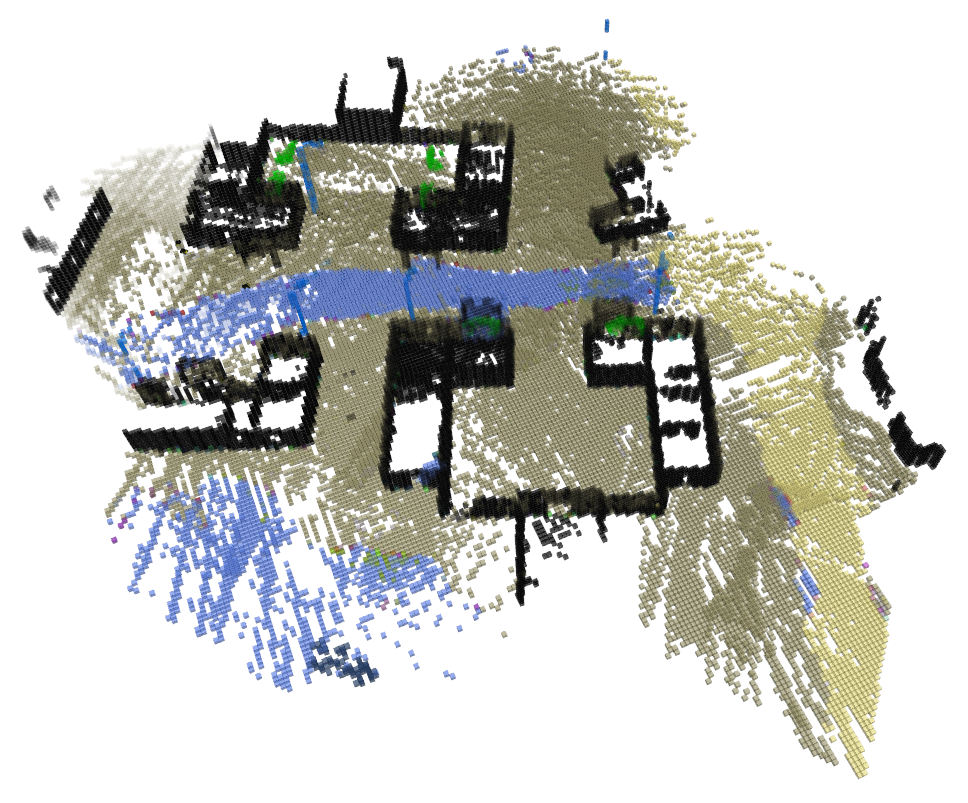}
    \captionsetup{justification=centering}
    \caption{Semantic octree map after $300$ s of exploration}
    \end{subfigure}
    \begin{subfigure}[t]{\linewidth}
    \centering
    \includegraphics[width=0.85\linewidth]{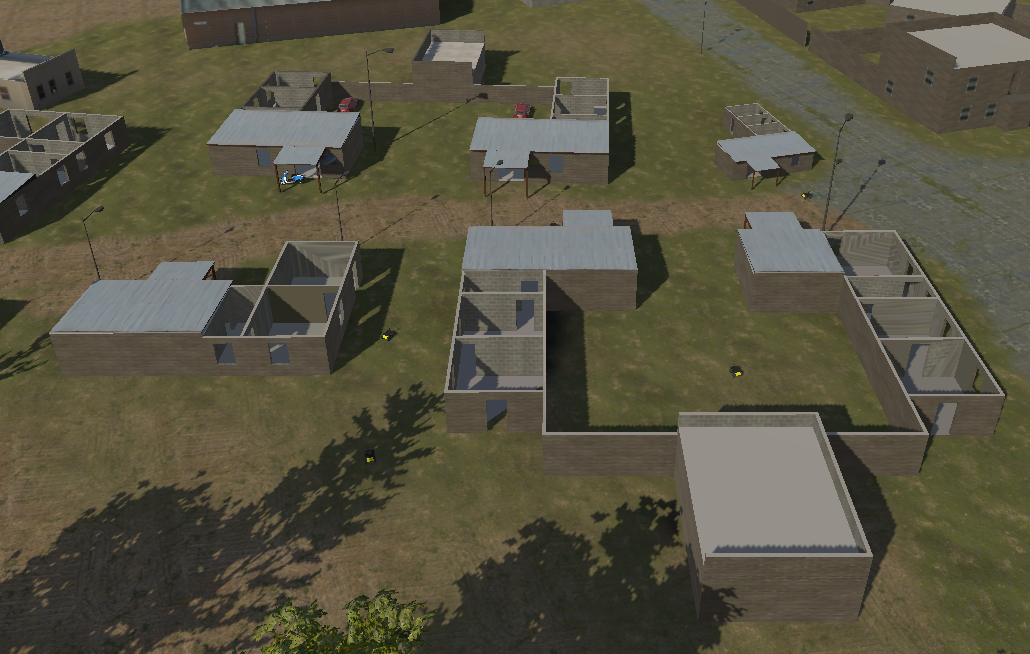}
    \captionsetup{justification=centering}
    \caption{Photo-realistic Unity simulation environment}
    \end{subfigure}
    \caption{Time lapse of multi-robot autonomous exploration and semantic octree mapping in the environment shown in (d). Different colors represent different semantic categories (building, dirt road, grass, etc.). Local semantic maps are overlaid with each other such that the transparency of each cell directly relates to the deviation between the local map estimates.}
    \label{fig:mapping}
\end{figure}

\begin{figure}[t]
    \centering
    \includegraphics[width=0.9\linewidth]{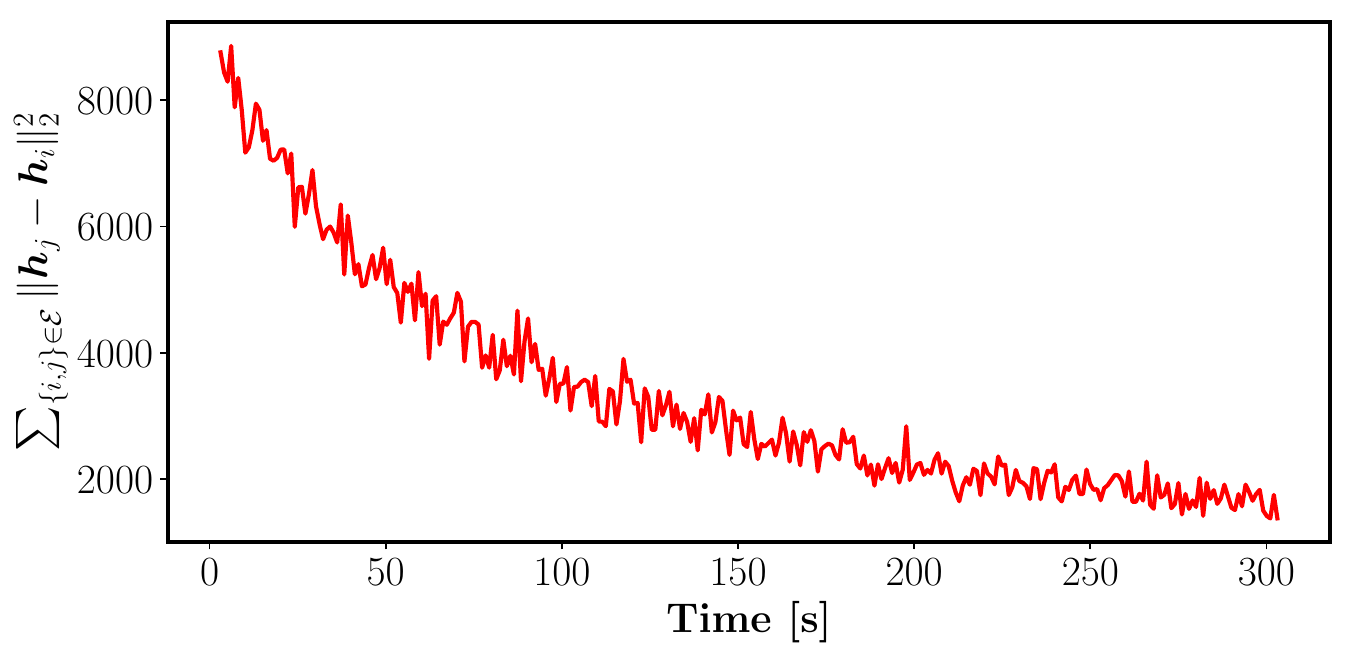}
    \caption{Time evolution of the total distance in local map estimates. Iteration of Alg.~\ref{alg:dist_mapping} is executed in each robot every one second, upon arrival of a local map estimate from a neighboring robot.}
    \label{fig:phi}
\end{figure}

\begin{figure}[t]
    \centering
    \includegraphics[width=0.9\linewidth]{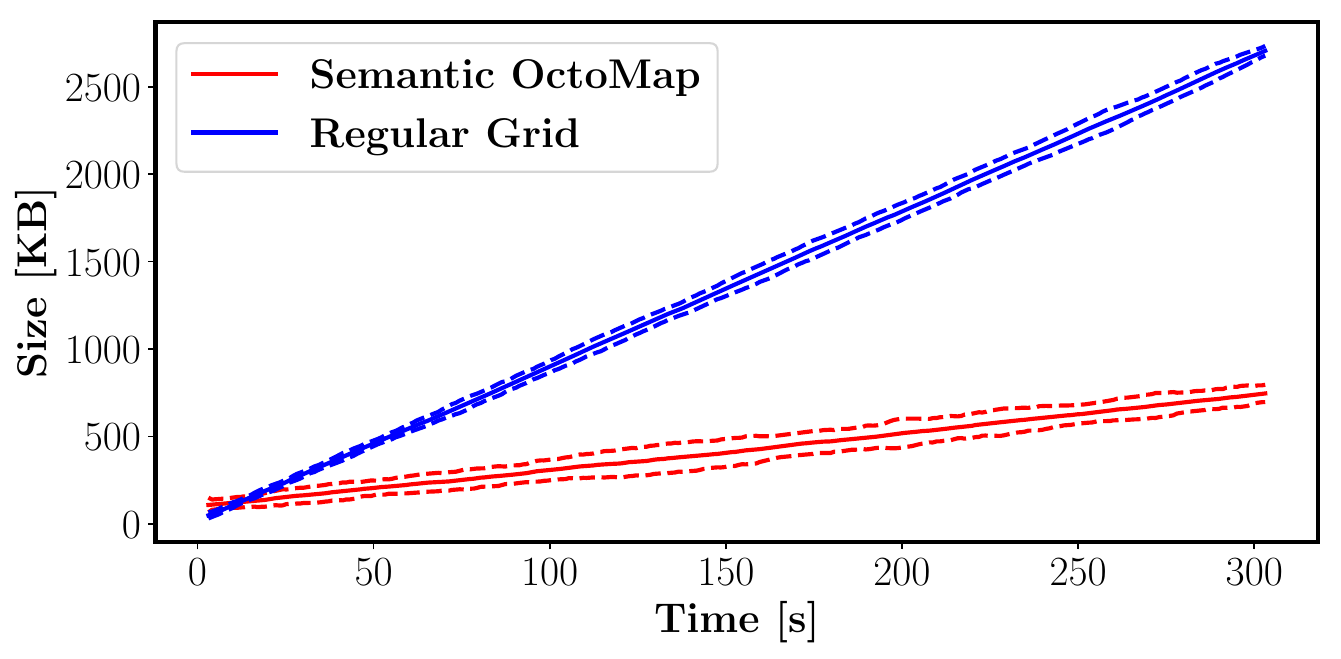}
    \caption{Packet size used for local map communication for uniform resolution grid map (blue) and semantic octree map (red). The solid lines and the dashed lines show the average over all robots and one standard deviation from the average, respectively.}
    \label{fig:comm_size}
\end{figure}

We tested our distributed semantic octree mapping algorithm in a 3-D photo-realistic Unity simulation environment, resembling an outdoor village area with various object classes, such as grass, dirt road, building, car, etc. Six ClearPath Husky robots each equipped with an RGB-D camera explore the unknown environment using frontier-based exploration \cite{frontier}. The RGB sensor measurements are used to perform pixel classification, which combined with the depth input, are provided as local observations to build semantic octree maps. We assume known robot poses and fully connected communication graph $\calG$ throughout the experiment. Each robot publishes its local map every $1$s, and upon arrival of a map, one iteration of Alg.~\ref{alg:dist_mapping} is executed.
Fig.~\ref{fig:mapping} shows several snapshots of the multi-robot semantic octree mapping, as well as the simulation environment. Fig.~\ref{fig:phi} shows the total distance in local map estimates, i.e. $\sum_{\{i,j\} \in \calE} \|\bfh_{j} - \bfh_{i}\|_2^2$, summed over all semantic octree map leaf nodes. Note that we did not account for the communication strength coefficients $A_{ij} \leq 1$ since doing so would underestimate the actual divergence in the local map estimates. As the figure shows, our distributed mapping algorithm leads to reduction of the total distance in local map estimates. Regarding communication requirements of our distributed semantic octree mapping, Fig.~\ref{fig:comm_size} illustrates the size of the communicated local semantic octree maps, with comparison to running the same distributed mapping experiment using uniform resolution grid maps. Our semantic octree map shows more than four times saving in communication bandwidth, empirically validating the benefit of our map representation for distributed multi-robot applications.

In this work, we presented distributed construction of semantic octree maps of unknown environments using a team of robots. The global map can be accessed via query from any of the robots, eliminating the need for a central estimation unit. Such common map can be utilized by robots for decentralized planning. In a future work, we use our multi-robot mapping algorithm alongside a distributed planning method in order to achieve multi-robot exploration.







{\small
\bibliographystyle{cls/IEEEtran}
\bibliography{bib/IEEEexample.bib}
}

\end{document}